\documentclass[letterpaper, 10 pt, conference]{ieeeconf}  

\IEEEoverridecommandlockouts                              

\usepackage{graphics} 
\usepackage{graphicx}
\usepackage{epsfig} 
\usepackage{mathptmx} 
\usepackage{times} 
\usepackage{amsmath} 
\usepackage{amssymb}  
\usepackage{booktabs}
\usepackage{makecell}
\usepackage{xcolor} 
\usepackage{multirow}
\usepackage{amssymb}
\usepackage{pifont}
\usepackage{listings}
\usepackage{wrapfig}
\usepackage{float}
\usepackage{ragged2e}

\usepackage{tabularx} 

\newcommand\our{RoboPro}
\newcommand\ourdata{Video2Code}

\title{\LARGE \bf
Robotic Programmer: Video Instructed Policy Code Generation for Robotic Manipulation
}

\author{Senwei Xie, Hongyu Wang, Zhanqi Xiao, Ruiping Wang and Xilin Chen
\thanks{*The authors are with the Key Laboratory of AI Safety of CAS, Institute of Computing Technology, Chinese Academy of Sciences (CAS), Beijing, 100190, China, and also with the University of Chinese Academy of Sciences, Beijing 100049, China. \tt\footnotesize\justifying{\{senwei.xie,hongyu.wang,zhanqi.xiao\}@vipl.ict.ac.cn, \{wangruiping,xlchen\}@ict.ac.cn}}}

\begin{document}

\maketitle
\thispagestyle{empty}
\pagestyle{empty}

\begin{abstract}

Zero-shot generalization across various robots, tasks and environments remains a significant challenge in robotic manipulation. Policy code generation methods use executable code to connect high-level task descriptions and low-level action sequences, leveraging the generalization capabilities of large language models and atomic skill libraries. In this work, we propose \textbf{Robo}tic \textbf{Pro}grammer (\textbf{\our{}}), a robotic foundation model, enabling the capability of perceiving visual information and following free-form instructions to perform robotic manipulation with policy code in a zero-shot manner. To address low efficiency and high cost in collecting runtime code data for robotic tasks, we devise \ourdata{} to synthesize executable code from extensive videos in-the-wild with off-the-shelf vision-language model and code-domain large language model. Extensive experiments show that \our{} achieves the state-of-the-art zero-shot performance on robotic manipulation in both simulators and real-world environments. Specifically, the zero-shot success rate of \our{} on RLBench surpasses the state-of-the-art model GPT-4o by 11.6\%, which is even comparable to a strong supervised training baseline. Furthermore, \our{} is robust to variations on API formats and skill sets. Our website can be found at \textcolor{magenta}{https://video2code.github.io/RoboPro-website/}.

\end{abstract}

\section{INTRODUCTION}

A long-term goal of embodied intelligence research is to develop a single model capable of solving any task defined by the user. Recent years have witnessed a trend towards large-scale foundation models on natural language processing tasks~\cite{gpt4, llama}. Scaling up these language models in terms of model size and training tokens significantly improves the few-shot performance on a range of end tasks, even achieving performance comparable to previous state-of-the-art fine-tuning methods. However, for robotic tasks, we have yet to see large-scale pre-trained models that can directly transfer across different robots, tasks and environments without additional fine-tuning.

To improve the zero-shot generalization ability of robotic models, one common approach is to unify different tasks as the next action prediction. This paradigm requires the model to directly generate low-level action sequences. \cite{rt2,rt-x, openvla} collected large amount of trajectories across various robots, tasks and environments. They trained vision-language-action (VLA) models derived from LLMs to map images and task instructions into discrete action tokens. Despite these models achieve better performance and show the capacity to transfer on novel objects and different tasks, fine-tuning is still required when deploying on new robots and environments. Besides, it is extremely expensive to collect trajectories through real-world robots, while using human-built simulators often leads to lack of diversity and introduces additional gap between simulation platform and real-world usages. 

Another line of research aims to use code as compromise solution for bridging high-level instructions and low-level robot execution, leveraging the generalization capabilities of Large Language Models (LLMs) and atomic skill libraries. RoboCodeX~\cite{robocodex} utilizes large vision-language model (VLM) to generate tree-of-thought plans and grasp preference. However, it also relies on manually-built simulation environment and human-annotated code for data curation, which is expensive and not friendly for scaling up in terms of training data.

In this work, we introduce \textbf{Robo}tic \textbf{Pro}grammer (\textbf{\our{}}), a robotic foundation model, enabling the capability of perceiving visual information and following free-form user instructions to perform manipulation tasks without additional fine-tuning. \our{} generates the executable code to connect high-level instructions and low-level action sequences. To address low efficiency and high cost in collecting runtime code data for robotic tasks, we devise \ourdata{}, an automatic data curation pipeline for multimodal code generation.

   \begin{figure*}[thpb]
      \centering
      \includegraphics[width=1\textwidth]{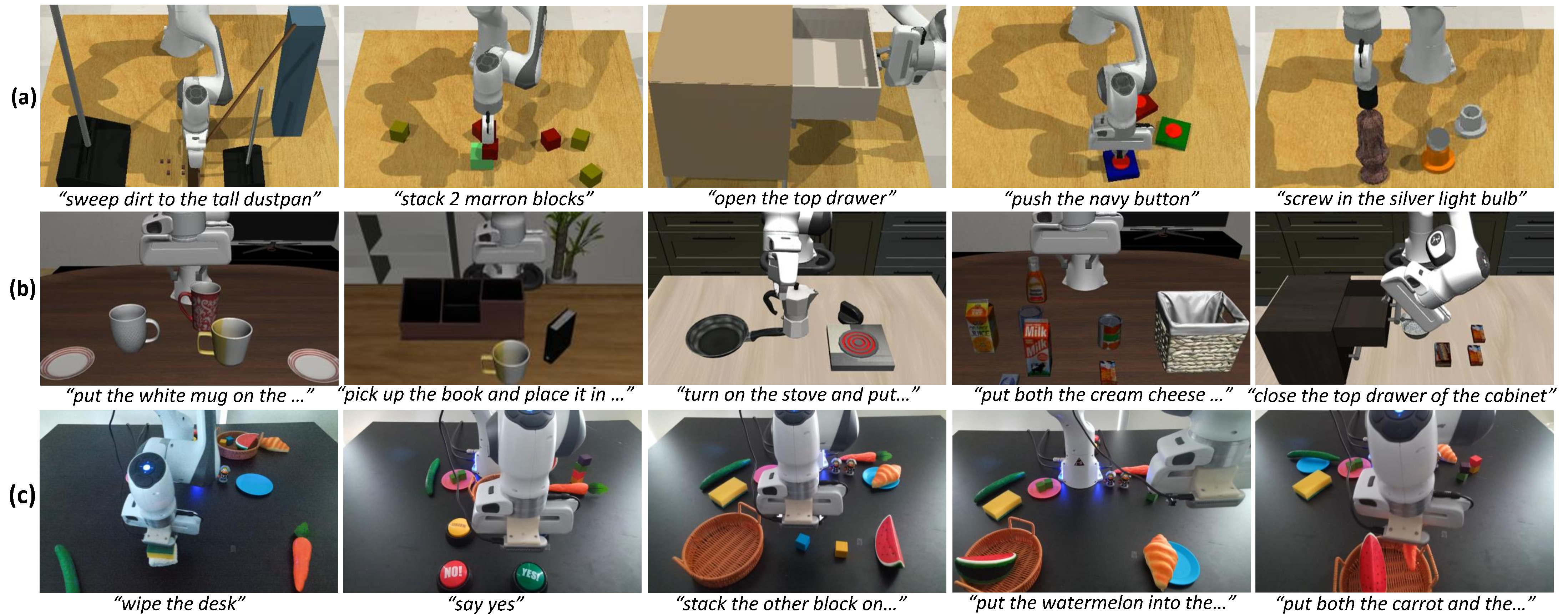}
      \caption{Visualization of evaluation tasks and execution results. \our{} shows impressive zero-shot performance on novel and compositional tasks in RLBench (a), long-termed manipulation tasks in LIBERO (b), and real-world tasks (c). Video demos can be found in our supplementary materials.}
      \label{fig:tasks}
   \end{figure*}

We draw our inspiration from the extensive amount of operational videos in-the-wild that implicitly contain necessary procedural knowledge about how to finish operational tasks. Previous research has focused on utilizing videos for large-scale supervised learning~\cite{rt2, openvla} or extracting relevant knowledge (e.g., affordance~\cite{bahl2023affordances}), while extracting executable policy code from videos is still under-explored. Our data curation pipeline uses the off-the-shelf VLM and Code LLM to synthesize code execution data from videos, which is much more efficient and scalable compared with generating code data from manually-built simulation environments. With \ourdata{}, we synthesize 115k robot execution code data along with the corresponding scene information and task descriptions from DROID~\cite{droid}. Extensive experiments (examples depicted in Fig.~\ref{fig:tasks}) show that \our{} achieves the state-of-the-art zero-shot performance on robotic manipulation tasks in both simulators and real-world environments. Specifically, the zero-shot success rate of \our{} on RLBench outperforms the state-of-the-art model GPT-4o by a gain of 11.6\%. It is even comparable to a strong supervised training method PerAct~\cite{peract}. Furthermore, we make an early attempt to discuss the adaptability of~\our{} across variations on API formats and unseen skill sets.

\section{Related Works}
\subsection{Language-Guided Robot Manipulation.} 
Language-conditioned robot manipulation refers to the use of natural language instructions to guide robotic actions. Natural language instructions allow non-experts to interact with robots through intuitive commands and enable robots to generalize to various tasks based on natural language input~\cite{winograd}. Recent advancements in language-conditioned embodied agents have leveraged Transformers~\cite{attention} to enhance performance on multi-task settings. One category of recent approaches is language-conditioned behavior cloning (BC), where models learn to mimic demonstrated language-conditioned actions and output dense action sequences directly. 3D BC methods~\cite{peract, sam-e} trained from scratch perform well on specific environment, while lacking of generalization ability across environments. Vision-language-action (VLA) models~\cite{rt2, openvla, llarva} built on pre-trained large language models (LLMs) show capacity to transfer on novel objects and task settings, but need additional fine-tuning when being deployed on new environments and robots. Another line is to create high-level planners based on LLMs~\cite{saycan, palme, inner}, which output step-by-step natural language plans according to human instructions and environmental information. These methods show better generalization ability across environments, leveraging the reasoning and generalization ability of LLMs on language instructions and environments. However, there is still a gap between generated natural language plans and low-level robotic execution, requiring an extra step to score potential actions or decompose plans into relevant policies~\cite{progprompt}.

\subsection{Robot-Centric Policy Code Generation.} Code-as-Policies~\cite{cap} proposes that executable code can serve as a more expressive way to bridge high-level task descriptions and low-level execution. Atomic skills to perceive 3D environments and plan primitive tasks are provided in predefined API libraries. LLMs process textual inputs and generate executable policy code conditioned on the API libraries~\cite{progprompt,cap,instruct2act,huang2023voxposer}. However, these methods rely solely on linguistic inputs, requiring detailed descriptions of environments and instructions as textual inputs, which limits their generalization and visual reasoning ability across environments. RoboCodeX~\cite{robocodex} utilizes large vision-language model (VLM) to decompose multimodal information into object-centric units in a tree-of-thought format. Nevertheless, it relies on manually-built simulation environments and human-annotated data, which lacks environmental richness and is expensive for scaling up. Different from previous works using language-only LLMs, \our{} enables visual reasoning ability and follows free-form instructions in a zero-shot manner. Furthermore, an automatic and scalable data curation pipeline \ourdata{} is developed to synthesize runtime code data from extensive videos in-the-wild in a quite efficient and low-cost fashion.

   \begin{figure*}[t]
      \centering
      \includegraphics[width=1\textwidth]{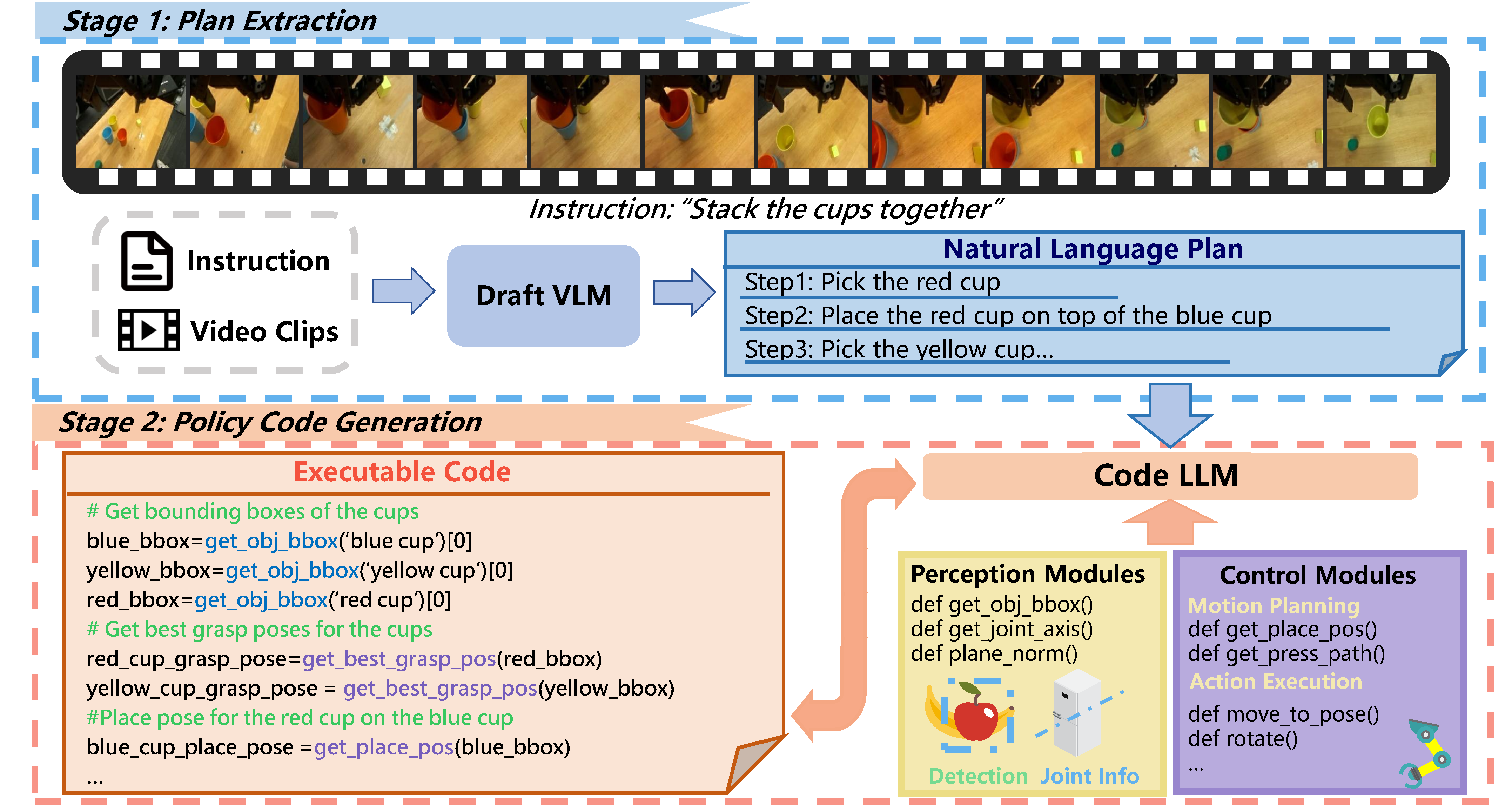}
    \caption{The data curation pipeline of \ourdata{}. We first use the Draft VLM to extract a brief natural language plan for execution of the user instruction. After that, the Code LLM generates robot-centric code using the provided API library and natural language plan from the first stage.}
    \label{fig:data}
   \end{figure*}

\section{Method}
\subsection{Problem Statement}
\label{section:Problem Statement}
We consider language-guided robotic manipulation where each task is described with a free-form language instruction $I$. Given RGBD data from the wrist camera as the observation space $O_t$ and robot low-dimension state $s_t$ (e.g., gripper pose at current time $t$), the central problem investigated in this work is how to generate motion trajectories $T$, where $T$ denotes a sequence of end-effector waypoints to be executed by an Operational Space Controller~\cite{OSC}. However, generating dense motion trajectories at once according to the free-form instruction $I$ is quite challenging, as $I$ can be arbitrarily long-horizon and would require comprehensive contextual understanding. Policy code generation methods map long-horizon instructions to a diverse set of atomic skills, leading to rapid adaptation capabilities across various robotic platforms. With comprehensive contextual understanding and advanced visual grounding capabilities, large vision-language models can function as intelligent planners, translating the task execution process into generated programs due to their robust emergent capabilities.

To prompt vision-language models (VLMs) to generate policy code, we assume a set of parameterized skills with unified interface, which is defined as the API library $L_{API}$. $L_{API}$ can be categorized into perception module $L_{per}$ and control module $L_{con}$ based on the API's role in task execution process. $L_{per}$ is tasked with segmenting the task-relevant part point cloud $\Pi_I$ and predicting the physical property $\phi_I$ of relevant objects, while $L_{con}$ predicts the contact pose of the gripper and generates the motion trajectory $T$ based on the output of $L_{per}$ and the current robot state $s_t$:

\begin{equation}
L_{API} = \{L_{per}, L_{con}\} 
\end{equation}
\begin{equation}
\{\Pi_I, \phi_I\} = L_{per}(O_t, I) 
\end{equation}
\begin{equation}
T = L_{con}(s_t, \{\Pi_I, \phi_I\}).    
\end{equation}

With the visual observation and the language instruction, VLMs generate executable policy code $\{\pi_i, p_i\}_{i=1}^N$ conditioned on the API library $L_{API}$, where $\pi_i$ denotes the $i$-th $L_{per}$ or $L_{con}$ calls and $p_i$ represents corresponding parameters for API calls. Each API call generates a sub-trajectory sequence $\tau_{i}$ of arbitrary length (the length is $\geq$ 0). All sub-trajectory sequences $\{\tau_{i}\}_{i=1}^N$ are then concatenated to form the final complete motion trajectory $T$. The whole generation process is formulated as:

\begin{equation}
	(O_t, I) \overset{VLM}{\Longrightarrow} \{\pi_i, p_i\}_{i=1}^N \Longrightarrow \{\tau_i\}_{i=1}^N.
\end{equation}

Explainable API calls generated by VLMs connect the observation and high-level instructions to low-level execution, enabling the capacity of zero-shot generalization in free-form language instructions and across different environments. Obviously, training such VLMs to perceive environments, follow instructions and generate executable code will inevitably require a vast amount of diverse and well-aligned robot-centric multimodal runtime code data, which poses a significant challenge. 

   \begin{figure*}[t]
      \centering
      \includegraphics[width=1\textwidth]{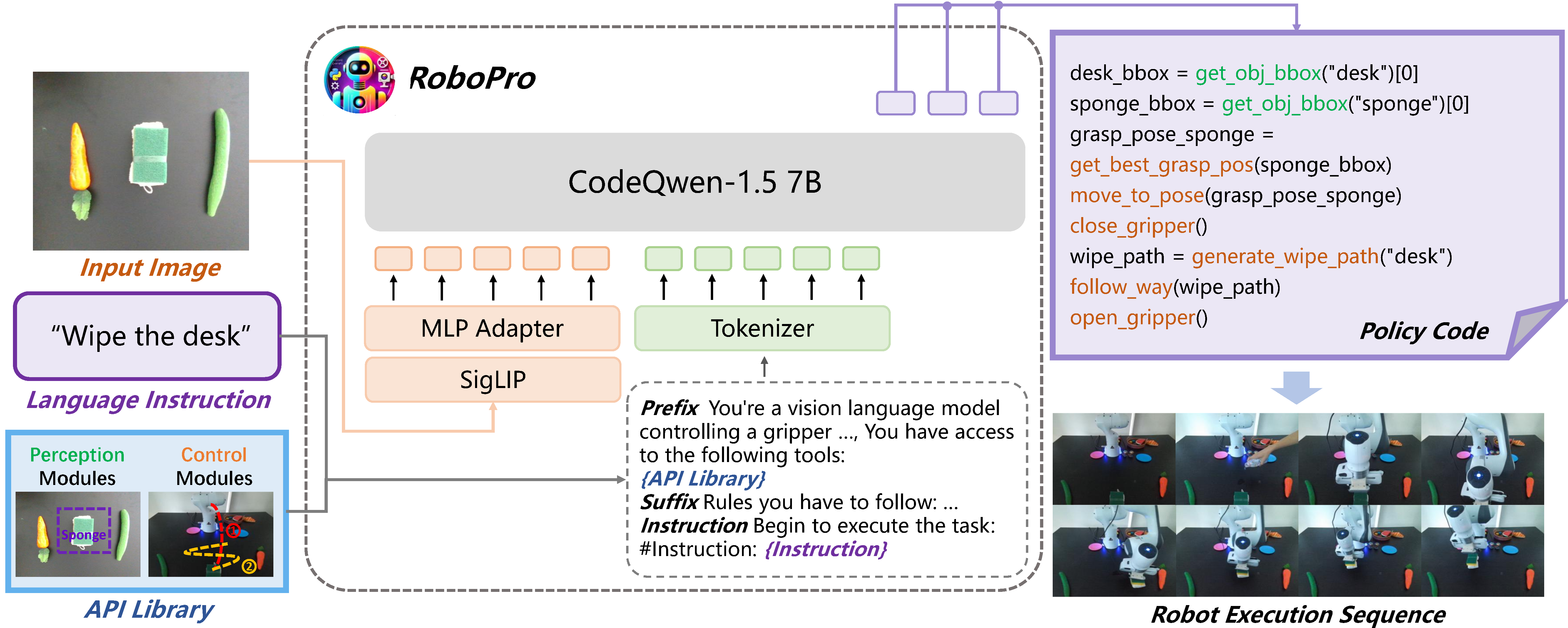}
    \caption{The overview of \our{}. \our{} utilizes environmental observation and natural language instruction as multimodal input, then outputs executable policy code. Extendable API library plays a role in mapping policy code into low-level execution sequences.}
    \label{fig:overall}
   \end{figure*}

\subsection{\ourdata{}: Synthesize Runtime Code From Videos}
\label{mt:video2code}
Videos are widely available raw data sources for runtime code data synthesis. Extensive operational videos naturally provide low-level details of performing tasks such as \textit{"how to pour tea into a cup"}, which inherently contain necessary procedural knowledge for runtime code data. Despite their favorable diversity and considerable quantity, it is still an under-explored and challenging problem how to collect executable policy code from demonstration videos efficiently. To this end, we devise \ourdata{}, a low-cost and automatic data curation pipeline to synthesize high-quality runtime code data from videos in an efficient way. Although open-source or lightweight vision-language models exhibit promising performance on video understanding tasks, a performance gap remains when compared to code-domain large language models in handling complex code generation tasks. As depicted in Fig.~\ref{fig:data}, to combine the visual reasoning ability of VLM and coding proficiency of code-domain LLM, \ourdata{} adopts a two-stage strategy. 

\textbf{Plan extraction.} 
The first stage is to extract robot-centric plans in natural language from instructional videos. These instructional videos are filtered from DROID~\cite{droid}, a large-scale robot manipulation dataset with 350 hours of interaction data across 564 scenes, 86 tasks, and 52 buildings. We extract 50k independent instructional videos with at least one free-form human instruction and further clip each video into 16 key frames. After that, we use Gemini-1.5-Flash~\cite{gemini} as the Draft VLM to generate a brief list of actions for human instruction with these key frames as reference. As shown in Fig.~\ref{fig:data}, the Draft VLM generates a step-by-step robot-centric plan from an instructional video to \textit{"stack the cups together"}. The generated natural language plans contain knowledge and habit of human to follow free-form embodied instructions, and key visual information is extracted automatically from the instructional video. 

\textbf{Policy code generation.} 
After plan extraction, we use Code LLM DeepSeek-Coder-V2~\cite{deepseek-v2} to "translate" these natural language plans into executable code. A complete prompt fed into the Code LLM includes API definitions, the natural language plan, and auxiliary part containing rules to follow. In the API definitions part, parameterized API functions are classified into two categories as formulated in Sec.~\ref{section:Problem Statement}: perception module, and control module. For each of these API functions, we provide API definitions and descriptions to demonstrate their usage. Auxiliary part contains prefix, third party tools, and rules to follow, similar to previous practices in RoboCodeX~\cite{robocodex}. Natural language plans accompanied with original human instructions are attached at the end of the prompt. As shown in Fig.~\ref{fig:data}, step-by-step decomposed natural language plan guides the Code LLM to generate high-quality policy code in a Chain-of-Thought format. As for API implementation, we use GroundingDINO~\cite{liu2023grounding} and AnyGrasp~\cite{fang2023anygrasp} to get the bounding boxes and grasp preferences, respectively. Besides, we provide heuristic implementation for compositional skills. We finally collect 115k runtime code data with task descriptions and environmental observations using \ourdata{} for supervised fine-tuning.

\subsection{\our{}: Robotic Foundation Model}
\label{mt:robopro}

\our{} introduces a unified architecture that seamlessly integrates visual perception, instruction following, and code generation by leveraging end-to-end vision-language models (VLMs). The unified pipeline eliminates the potential loss of critical information during intermediate steps and enhances computational efficiency during inference. Powered by well-aligned image-instruction-code pairs from~\ourdata{}, ~\our{} demonstrates strong capabilities in executing free-form instructions grounded in visual observations.

\textbf{Model architecture.} 
As shown in Fig.~\ref{fig:overall}, \our{} has a vision encoder and a pre-trained LLM. They are connected with a lightweight adaptor layer consisting of a two-layer MLP. Specifically, the vision backbone first encodes the image into a sequence of visual tokens. After that, the lightweight adaptor is designed to project visual tokens onto embedding space of the LLM. In addition, we provide the API definitions and the user instruction as the text inputs. The visual and text tokens are directly concatenated and then fed into the LLM. The LLM are trained to generate the runtime code based on the visual inputs and task description. 

\our{} is designed to reason on multimodal inputs and generate executable policy code for robotic manipulation. Thus, two key factors for the choice of its components are the ability of visual reasoning and the quality of policy code generation. \our{} adopts SigLIP-L~\cite{sigmoid} as the vision encoder, which yields favorable performance on general visual reasoning tasks. For the base LLM, a code-domain LLM, CodeQwen-1.5~\cite{qwen}, is utilized, which shows state-of-the-art performance among open-source code models. The model architecture and working process of \our{} are illustrated in Fig.~\ref{fig:overall}.

\textbf{Training.} 
The training procedure of \our{} consists of three stages: visual alignment, pre-training, and supervised fine-tuning (SFT). We first train a lightweight adaptor layer while freezing the vision encoder and LLM with LLaVA-Pretrain~\cite{llava1_5}. Then we pre-train the lightweight adaptor and the LLM on a corpus of high-quality image-text pairs~\cite{sharegpt4v}. For supervised fine-tuning, the 115k runtime code data generated by \ourdata{} (as noted in Sec.~\ref{mt:video2code}) are used. To avoid overfitting and enhance visual reasoning ability, a general vision language fine-tuning dataset (LLaVA-1.5~\cite{llava1_5}) is also involved during the SFT process. Thus, \our{} is trained to follow free-form language instructions and perceive visual information to generate executable policy code for robotic manipulation. 

\begin{table*}[t]
    \caption{Success rate (\%) on RLBench Multi-Task setting. PerAct greyed on need supervised training on the simulation platform.}
    \label{tab:rlbench}
    \begin{center}
    \begin{tabular}{l|ccccccccccc}
    \toprule
    \bf Models &  \makecell{Push\\Buttons} &  \makecell{Stack\\Blocks} &  \makecell{Open\\Drawer} &  \makecell{Close\\Jar}  &  \makecell{Stack\\Cups} &  \makecell{Sweep\\Dirt} &  \makecell{Slide\\Block} &  \makecell{Screw\\Bulb} &  \makecell{Put in\\Board} &  Avg. \\
    \midrule 
    \color{gray!70} PerAct~\cite{peract} & \color{gray!70} 48 & \color{gray!70} 36 & \color{gray!70} 80 & \color{gray!70} 60 & \color{gray!70} 0 & \color{gray!70} 56 & \color{gray!70} 72 & \color{gray!70} 24 & \color{gray!70} 16 &  \color{gray!70} 43.6\\
    CaP~\cite{cap} & \bf 72 & 4 & 24 & 40 & 0 & 36 & 4 & 20 & \bf 12 & 23.6\\
    GPT-4o~\cite{gpt4o} & \bf 72 & 20 & 56 & 36 & 4 & 40 & 20 & 20 & \bf 12 & 31.1\\
    \midrule
    \bf \our{} (ours) & 68 & \bf 48 & 68 & 44 & 4 & \bf 48 & 60 & 32 & \bf 12 & \bf 42.7\\
    \quad w/ API Renaming & 68 & 40 & 60 & \bf 48 & 4 & \bf 48 & 68 & \bf 36 & \bf 12 & \bf 42.7\\
    \quad w/ API Refactoring & 68 & 36 & \bf 72 & 44 & \bf 8 & 16 & \bf 80 & 28 & \bf 12 & 40.4\\
    \bottomrule
    \end{tabular}
    \end{center}
\end{table*}

\begin{table*}[t]
    \caption{Success rate (\%) on 8 tasks on LIBERO. OpenVLA greyed on is fine-tuned on this simulation platform.}
    \label{tab:libero}
    \begin{center}
    \begin{tabular}{l|ccccccccc}
    \toprule
    \bf Models &  \makecell{Turn on\\Stove} &  \makecell{Close\\Cabinet} &  \makecell{Put in\\Sauce} &  \makecell{Put in\\Butter}  &  \makecell{Put in\\Cheese} &  \makecell{Place\\Book} &  \makecell{Boil\\Water} &  \makecell{Identify\\Plate} &  Avg. \\
    \midrule 
    \color{gray!70} OpenVLA~\cite{openvla} & \color{gray!70} 97 & \color{gray!70} 97 & \color{gray!70} 37 & \color{gray!70} 60 & \color{gray!70} 53 & \color{gray!70} 93 & \color{gray!70} 43 & \color{gray!70} 40 & \color{gray!70} 65.0\\
    CaP~\cite{cap} & 0 & 37 & 17 & 13 & 7 & 30 & 7 & 7 & 14.8\\
    GPT-4o~\cite{gpt4o} & 37 & 17 & 63 & 43 & 57 & \bf 43 & 17 & 3 & 35.0\\
    \midrule
    \bf \our{} (ours) & \bf 97 & \bf 60 & \bf 67 & \bf 53 & \bf 63 & \bf 43 & \bf 23 & \bf 13 & \bf 52.4\\
    \bottomrule
    \end{tabular}
    \end{center}
\end{table*}
   
\section{Experiments}
\label{section:exp}
\subsection{Zero-Shot Generalization across Tasks and Environments}
\label{section:exp_rlbench}

Zero-shot generalization across instructions, tasks and environments is a significant challenge for robotic learning. Policy code generation methods leverage adaptability of large language models to generate code plans across tasks and scenarios in a zero-shot manner.~\our{}, a multimodal policy code generation model, is trained on real-world data. To validate zero-shot generalization of~\our{} across environments, we evaluate the performance on two distinct simulation environments (RLBench, LIBERO), and carefully verified that scenes, tasks and instructions during testing were entirely unseen during the training phase.

The baselines can be categorized into two groups. Our primary comparison tagets are other code generation methods. They first output robot-centric policy code, then execute it with provided APIs. We evaluate their zero-shot performance on RLBench and LIBERO. CaP~\cite{cap} equips large language model with the ground-truth textual scene descriptions, containing object names, attributes, and instructions, to generate executable code. Following their paper, we implement CaP with GPT-3.5-Turbo (gpt-3.5-turbo-0125). GPT-4o (\cite{gpt4o}, gpt-4o-2024-05-13) is the state-of-the-art multimodal model for various vision-language tasks. For \our{} and GPT-4o, we require the model to directly generate the executable code given the image from the wrist camera, user instructions and API definitions. For a fair comparison, we adopt the same API library for these methods (i.e., CaP, GPT-4o, and \our{}). Our API library shares similar design formulation as RoboCodeX~\cite{robocodex}, with detailed implementation in Appendix~\ref{ap:prompt}. The methods from another group directly output actions while require supervised training when implement on different tasks and environments, e.g., behavior cloning methods, including PerAct~\cite{peract} and OpenVLA~\cite{openvla}. We use popular training-based methods primarily as an upper bound for reference.

\subsubsection{RLBench}
Following PerAct~\cite{peract}, we select 9 tasks with the requirement of novel instruction understanding or long-horizon reasoning in RLBench~\cite{rlbench} for evaluation. Each task is evaluated with 25 episodes scored either 0 or 100 for failure or success in task execution. Code generation planning methods are evaluated in a zero-shot manner with same API implementation. We use PerAct trained on 100 episodes in a multi-task setting for a training-based reference on this benchmark. Detailed experiment settings and task information in RLBench can be found in Appendix~\ref{ap:task_rlbench}.

\begin{figure}[thpb]
      \centering
      \includegraphics[width=0.45\textwidth]{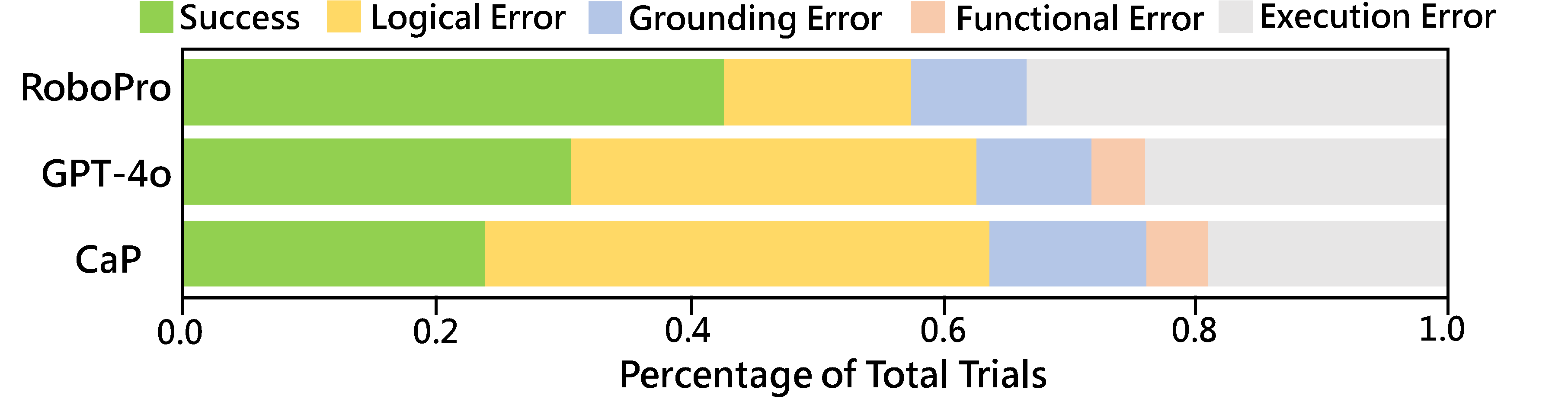}
      \caption{Error breakdown on RLBench.}
      \label{fig:error_breakdown}
   \end{figure}

We report the average success rate on 25 episodes for each task. As shown in Table~\ref{tab:rlbench}, the zero-shot result of \our{} surpasses language-only policy code generation method (CaP) by 19.1\%. Besides, our model significantly outperforms the state-of-the-art VLM GPT-4o by 11.6\% on average success rate. To thoroughly analyze the factors contributing to the performance gap between different methods, we conducted an error breakdown for the policy code generation approaches. In the context of policy code generation methods, the successful execution of manipulation tasks relies on both the accuracy of the policy code and the capabilities of the API library. The main types of errors impacting the quality of robot-centric policy code are logical errors, functional errors, and grounding errors. These errors are associated with challenges in the appropriate selection and utility of APIs, as well as issues related to visual grounding. As depicted in Fig.~\ref{fig:error_breakdown}, the results show that all these methods perform well on following functional definition of API library, causing a low occupancy of functional error. Compared with linguistic only method CaP, GPT-4o and \our{} show a noticeable improvement in target object grounding. The main failure cases of CaP and GPT-4o fall in logical error, including API selection and proper order of API calls. In contrast, \our{} effectively reduces this margin, mainly owing to the procedural knowledge about long-term execution learned in \ourdata{}. Execution errors maintain a consistent proportional relationship with successful cases, which result from API limitations rather than inaccuracies in the policy code. 

\subsubsection{LIBERO}
We choose 8 representative tasks from LIBERO~\cite{libero} as the evaluation set. These tasks include short-horizon tasks which need scene understanding, and long-horizon tasks which require multi-step implementation. Similar with RLBench, each task is evaluated with 30 episodes scored either 0 or 100 for failure or success execution. Detailed task descriptions and corresponding examples can be found in Appendix~\ref{ap:task_libero}. We fine-tuned OpenVLA-7B~\cite{openvla} using Low-Rank Adaptation (LoRA) across 8 evaluation tasks as reference of training-based methods. Code generation approaches keep zero-shot setting during evaluation. 

As reported in Table~\ref{tab:libero}, \our{} significantly outperforms GPT-4o by a gain of 17.4\% average success rate on 8 LIBERO tasks, which is aligned with the observations from the experiments on RLBench. Compared with GPT-4o, \our{} executes more accurate sequences of actions to complete various manipulation tasks. For instance, when given the task \textit{"Turn on the stove"}, \our{} consistently approaches the stove knob, grasps it, and rotates it clockwise. In contrast, GPT-4o sometimes misinterprets the knob's affordance, attempting to press it rather than rotate.

\subsection{Zero-Shot Generalization across APIs and Skills}
Another challenge for code generation methods lies in generalization across different robotic configurations, which often manifests as variations in the format and implementation of API libraries. Additionally, real-world tasks may require robots to perform unseen skills with user-specific preferences. Despite this issue has been largely under-explored in prior works, enhancing adaptability to diverse API implementations and novel skills is crucial for making code generation methods more scalable in real-world applications. We take an early step toward examining the robustness of code generation methods across varying API formats and user-specified skills.

\subsubsection{Generalization across different API formats}
\label{exp:api}
We first evaluate the robustness of \our{} to the changes of API formats, which implies that the model can understand and internalize the atomic skills under the API interface. To assess the generalization of \our{} under different level of changes in API library, we designed two representative sets of experiments: the \textit{API Renaming} set and the \textit{API Refactoring} set. For renamed APIs, we change in their names and keep consistent in functional structure (e.g., the type of return values and arguments). For refactored APIs, we change in functional structure but keep their names. Take the control API "\texttt{get\_best\_grasp\_pose()}" as an example. In the API Renaming set, it is renamed as "\texttt{generate\_obj\_grasp\_pos()}" without changes on functionality, and in the API Refactoring set, the inputs, outputs and comments are all changed (e.g., the input format changes from "\texttt{bbox}" to "\texttt{np.ndarray}"). As shown in Table~\ref{tab:rlbench}, the performance of \our{} on RLBench is robust to the changes in API formation, which originates from \our{}'s ability as a generalist code model to comprehend different variations of API formats. The detailed implementations of renamed and refactored APIs can be found in Appendix~\ref{ap:prompt}. 

\begin{table}[t]
    \caption{Success rate (\%) on three compositional tasks from RLBench based on a new set of task-Specific APIs.}
    \label{tab:skill}
    \begin{center}
    \begin{tabular}{l|cccc}
    \toprule
    \bf Models &  Water Plants &  Hit Ball &  Scoop Cube &  Avg. \\
    \midrule 
    CaP~\cite{cap} & 4 & 16 & 0 & 6.7 \\
    GPT-4o~\cite{gpt4o} & 40 & 12 & 24 & 25.3\\
    \midrule
    \bf \our{} (ours) & \bf 40 & \bf 44 & \bf 48 & \bf 44.0\\
    \bottomrule
    \end{tabular}
    \end{center}
\end{table}

\subsubsection{Generalization across unseen skills}

To further evaluate \our{}'s adaptability to newly defined or task-specific APIs, we select three compositional tasks from RLBench that involve multi-step execution: Water Plants, Hit Ball, and Scoop Cube. For each task, we design a new set of task-specific APIs encompassing skills not included in RoboPro's training phase, while follow similar functional structure with the original implementation of action modules. Under this setting, the performance of \our{} consistently outperforms CaP and GPT-4o in a zero-shot manner. We observe that on unseen skill sets, \our{} exhibits preferences and behaviors similar to those observed in the training skill set. Compared with methods using proprietary models, \our{} trained with video demonstrations tends to grasp tools with appropriate affordance before performing compositional skills, and matches visual observation with vague language instructions to perform in a comprehensive manner. This robustness implies that sequential action knowledge and preferences learned from Video2Code is transferable and beneficial to perform with unseen skill sets.

\begin{table}[t]
\centering
\caption{The zero-shot success rate (\%) of \our{} and GPT-4o across 8 real-world manipulation tasks.}
    \label{tab:real}
    \begin{tabular}{l|cccc}
    \toprule
    \bf Task &  \bf GPT-4o & \bf \our{} (ours) \\
    \midrule 
    Move in Direction  & 60 &  80  \\
    Setup Food & 80 &  90 \\
    Distinct Base &   80 & 70 \\
    Prepare Meal & 60 &  60  \\
    Tidy Table & 40 &  70 \\
    Express Words & 50 &  60 \\
    Stack on Color  & 10 &  50  \\
    Wipe Desk & 100 & 100 \\
    \bf Average & 60.0 & \bf 72.5 \\
    \bottomrule
    \end{tabular}
\end{table}

\subsection{Real-World Experiments}
To evaluate the performance of \our{} in real-world scenarios, we conduct realistic experiments on a Franka Emika robot arm equipped with an Intel RealSense D435i wrist camera. As emphasized in Sec.~\ref{section:Problem Statement}, long-horizon task decomposition and visual understanding capabilities are crucial for zero-shot generalization in language-guided robotic manipulation. To assess \our{}'s performance in these aspects, we carefully designed 8 tasks, ranging from short-horizon to long-horizon tasks, as well as tasks that require visual comprehension. For instance, \our{} is required to select the object with "wipe" affordance from the scene given instruction \textit{"wipe the desk"}. Additionally, to rigorously validate \our{}'s generalization capability across different real-world scenarios, we ensure that each task consists of at least two variations in terms of object categories and physical properties (10 tests are run for each task). We select GPT-4o as an extra baseline on real-world environments. We also provide detailed real-world setup in Appendix~\ref{ap:real-world-setting}.

As shown in Table~\ref{tab:real}, \our{} is able to achieve 72.5\% success rate on average among all 8 tasks, which verifies \our{}'s strong generalization ability in real-world scenarios without any specific fine-tuning. We also observe \our{} exhibits impressive emergent ability in visual reasoning. For example, when asked to wipe the desk, \our{} will choose the appropriate tool (the sponge) among irrelevant objects, and grasp it to wipe water on the desk. On directional moving and one-turn tasks, GPT-4o shows comparable performance with RoboPro (Move in Direction, Setup Food), while RoboPro shows better performance on tasks requiring visual understanding or target identification (Stack on Color, Tidy Table), which yields conclusions consistent with the experimental results on simulation platforms.

   \begin{figure}[t]
      \centering
      \includegraphics[width=0.45\textwidth]{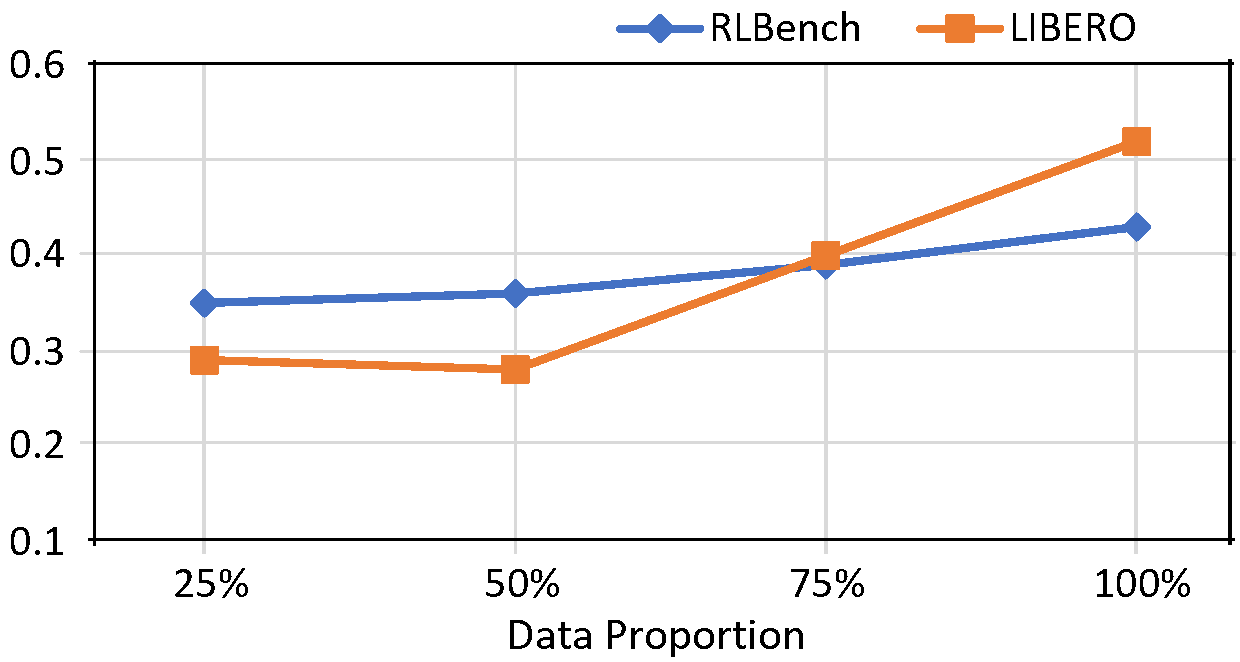}
    \caption{Success rate on manipulation tasks across varying data proportions.}
    \label{fig:scaling}
   \end{figure}

\begin{table}[t]
    \centering
    \small
    \caption{Ablations of \ourdata{} and different base LLMs.}
    \label{tab:ablation}
    \begin{tabular}{l|c|cc}
    \toprule
     \multirow{2}{*}{\bf LLM} & \multirow{2}{*}{\bf \ourdata{}} & \multicolumn{2}{c}{\bf Manipulation}  \\ 
     & & RLBench & LIBERO \\
    \midrule 
    CodeQwen-1.5-7B & \ding{55} & 0.4  & 7.0 \\
    CodeQwen-1.5-7B  & \ding{51} & \bf 42.7  & \bf 52.4 \\
    DeepSeek-Coder-6.7B & \ding{51} &  41.3  & 48.8 \\
    \bottomrule
    \end{tabular}
\end{table}

\subsection{Ablation Study}

We conduct extensive ablation studies to analyze how \ourdata{} and the scaling of training data impact performance on manipulation tasks. Additionally, we explore the effect of the choice of base LLM within \our{}. We provide detailed ablation results in Appendix~\ref{ap:exp}.

\textbf{Effectiveness of \ourdata{}.} We compare our model trained with and without \ourdata{} on manipulation tasks. For a fair comparison, we only remove \ourdata{} from the fine-tuning stage for the baseline, that is, the 115k runtime code data are excluded and only the general vision language fine-tuning dataset is used during the SFT process, as described in Sec.~\ref{mt:robopro}. The first two rows of Table~\ref{tab:ablation} show the comparison of the two settings. It is found that the \ourdata{} generated data have significantly improved the performance on both RLBench and LIBERO by a gain of 42.3\% and 45.4\%, respectively, which indicate \ourdata{}’s efficacy in enhancing the ability of skills utility and instruction following. 

\textbf{Scaling of training data.}  We further conducted an ablation study on the dataset size. Specifically, we trained RoboPro using 115k runtime code data collected from DROID, varying the dataset proportion of SFT stage to 25\%, 50\%, 75\%, and 100\%. We evaluated the models trained with different sizes of dataset on RLBench and LIBERO. As shown in Fig.~\ref{fig:scaling}, results indicate that RoboPro adheres to the scaling law: training with just 25\% of the data already yields a well-performing model, while its performance continues to improve as the dataset size increases. Video2Code is efficient for scaling up of runtime code data, which deserves further exploration to involve in more robotic demonstrations.

\textbf{Choice of base LLM.}
For the components of the \our{} framework, we evaluate its performance using different code-domain base LLMs, specifically DeepSeek-Coder-6.7B-Instruct~\cite{deepseek} and CodeQwen-1.5-7B-Chat~\cite{qwen}. As shown in Table~\ref{tab:ablation}, the version of \our{} trained with CodeQwen-1.5-7B-Chat consistently outperforms the one trained with DeepSeek-Coder-6.7B-Instruct across both manipulation tasks. These results demonstrate that employing a more powerful base LLM for code generation task can consequently enhance performance in both tasks.

\section{Conclusion and Future work}
In this work, we propose \our{}, a robotic foundation model, which perceives visual information and follows free-form instructions to perform robotic manipulation in a zero-shot manner. To address low efficiency and high cost for runtime code data synthesis, we propose Video2Code, a scalable and automatic data curation pipeline. Through extensive experiments, with assistance of \ourdata{}, \our{} achieves impressive generalization capability compared with training-based methods, and exhibits significant improvement on performance compared with other policy code generation methods. These results indicate that incorporating procedural knowledge within operational videos into training process will bring substantially enhanced understanding of skills (i.e., API libraries) and free-form instructions. Beyond the scope of robotic manipulation tasks, policy code generation methods also show potential in many other robotic applications (e.g., navigation). In the future, we would like to expand our method to more application scenarios to provide more comprehensive support for complex real-world robotic deployments. 

\section*{ACKNOWLEDGMENT}
This work is partially supported by National Key R\&D Program of China No. 2021ZD0111901,
 2023YFF1105104, and Natural Science Foundation of China under contract No. U21B2025. Senwei Xie, Hongyu Wang, and Zhanqi Xiao contributed equally to this work. Senwei Xie was primarily responsible for the implementation of Video2Code, evaluation on RLBench and real-world tasks. Hongyu Wang mainly provided support on the training of \our{}. Zhanqi Xiao  was primarily responsible for evaluation on LIBERO and setup of the real-world environment.

\bibliographystyle{IEEEtran}
\bibliography{IEEEexample}

\section*{APPENDIX}
\label{ap:task}
\subsection{Tasks in RLBench} 
\label{ap:task_rlbench}
RLBench is a simulation platform set in CoppelaSim~\cite{coppeliasim} and interfaced through PyRep~\cite{pyrep}. Robotic models control a 7-dof Franka Panda robot with a parallel gripper to complete language-conditioned tasks. \our{} is evaluated on 9 tasks from RLBench~\cite{rlbench}. Modification on these tasks is consistent with PerAct~\cite{peract}. Each task in RLBench is provided with several variations on language instructions describing the goal. In order to validate \our{}'s adaptation ability across various and vague instructions, we pop out an instruction from the language template list for each episode during evaluation instead of just using the first language template. Detailed descriptions and modification for each task in RLBench are provided below.

\paragraph{Push Buttons.} Push down colored buttons in a specific order. The task has 20 different variances on the color of buttons, and three variances on the number of buttons to be manipulated. The success metric of this task is to push down specific buttons in correct order.

\paragraph{Close Jar.} Put the lid on the table onto the jar with specific color. This task also has 20 different variations on the color of the jars. The success metric is that the lid is on the top of the target jar, and the gripper doesn't grasp anything.

\paragraph{Stack Blocks.} Stack two to four blocks with specific color onto the green target area. There are always two groups of four blocks with the same color, and this task has 20 variations on the color of the blocks. The success metric has a further requirement that all stacked blocks inside the area of a green platform beyond the original language instruction. We add target prompt to specify the stacking area.

\paragraph{Open Drawer.} Open specific drawer of a cabinet. there are three different variations on the position of the drawer: top, middle, and bottom. The success metric is a full extension of the target drawer joint. Before execution, we first adjust the gripper position to face the cabinet.

\paragraph{Stack Cups.} Stack other two cups onto the cup with specific color. This task has 20 variations on the color of the cups. The success metric of this task is that the other cups are inside the target cup. 

\paragraph{Sweep Dirt.} Sweep dirt particles to the target dustpan. There are two dustpans specified as a tall dustpan and a short dustpan. The success metric of this task is that all 5 dirt particles are in the target dustpan. This task is modified by PerAct. 

\paragraph{Slide Block.} Slide the red cube in the scene to the target colored area. There are four areas with different color on each corner of the scene, and the cube cannot be picked up. The success metric is that the cube is inside the area with the target color, which is modified by PerAct. 

\paragraph{Screw Bulb.} Screw light bulb with the specified base onto the lamp base. There are two bulbs in the scene at once, and the color of the holders have 20 different variations. The success metric is that the bulb is inside the lamp stand. 

\paragraph{Put in Board.} Pick up the specified object and place it into the cupboard above. There are always 9 different objects on the table. The success rate is that the target object is in the cupboard. 

\paragraph{Water Plants.} Pick up the watering can and pour water onto the plant. Five tiny cubes in the watering can represent the water. The success metric is that all water particles are in the area of the plant. A user specified API for this task is provided to move the watering can towards the plant and pour water into it.

\paragraph{Hit Ball.} Use the stick to hit the ball into the goal. The success metric is that the ball is in the target and the robot is grasping the stick. We design user specified API for this task to move the queue in front of the ball and hit it to the goal.

\paragraph{Scoop Cube.} Use the spatula to scoop the cube and lift it. The success metric is that the gripper is grasping the spatula, and the cube is in specific area on the top of its original position. We design user specified API for this task to scoop target and lift it to required height. 

\begin{table}[t]
    \centering
    \caption{The manipulation tasks selected for the evaluation of zero-shot generalization on LIBERO.}
    \label{tab:task_detail}
    \footnotesize 
    \begin{tabularx}{\linewidth}{l|X}
    \toprule
    \bf Task ID & \bf Task Instruction \\
    \midrule
    Turn on Stove  & turn on the stove \\
    Close Cabinet & close the top drawer of the cabinet \\
    Put in Sauce & put both the alphabet soup and the tomato sauce in the basket \\
    Put in Butter & put both the cream cheese box and the butter in the basket \\
    Put in Cheese & put both the alphabet soup and the cream cheese box in the basket \\
    Place Book & pick up the book and place it in the back compartment of the caddy \\
    Boil Water & turn on the stove and put the moka pot on it \\
    Identify Plate & put the white mug on the left plate and put the yellow and white mug on the right plate \\
    \bottomrule
    \end{tabularx}
\end{table}

\begin{figure*}[t]
    \centering
    \includegraphics[width=1\textwidth]{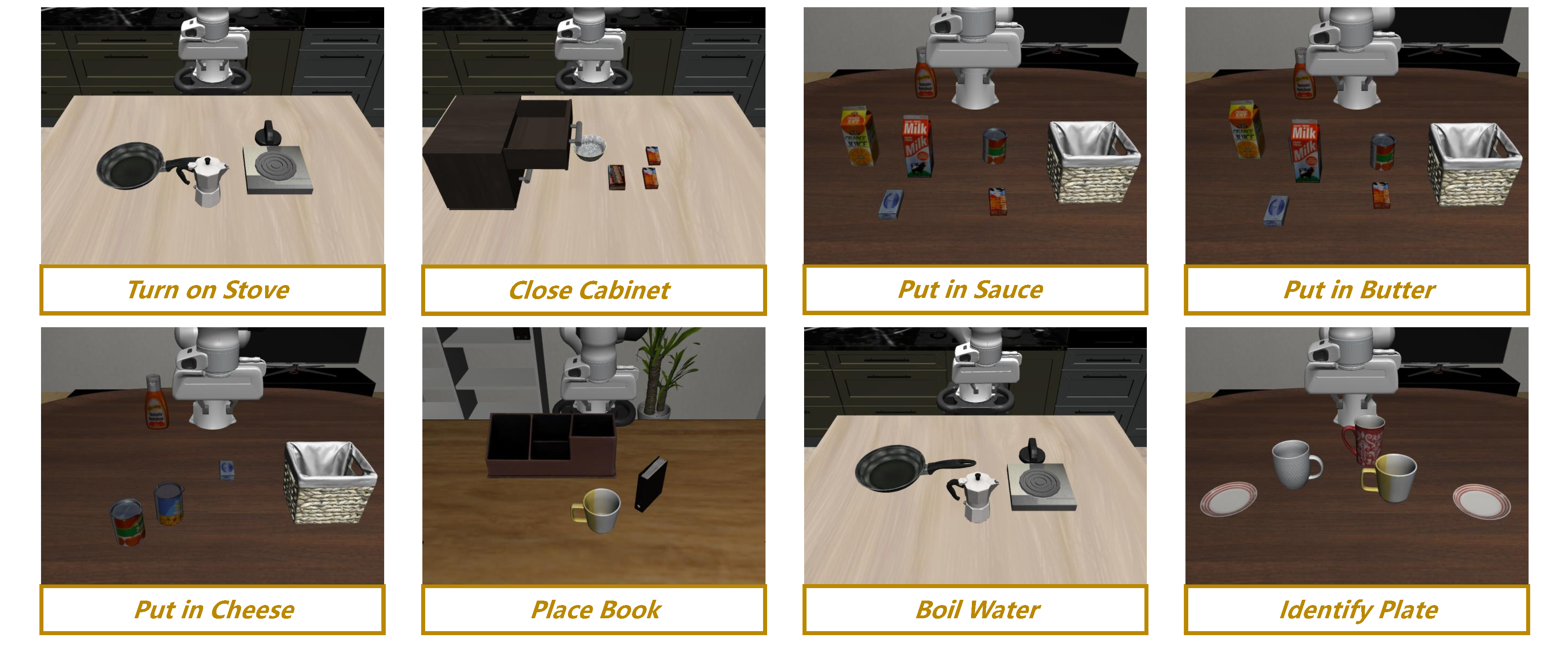}
    \caption{Illustration of the selected tasks from LIBERO benchmark.}
    \label{fig:libero_task}
\end{figure*}

\begin{table*}[t]
    \centering
    \caption{Detailed success rate (\%) of ablation study on the RLBench tasks. }
    \label{tab:ablation_rlbench}
    \small
    \setlength{\tabcolsep}{3pt} 
    \begin{tabular}{l|c|ccccccccccc}
    \toprule
    \bf LLM & \bf \ourdata{} &  \makecell{Push\\Buttons} &  \makecell{Stack\\Blocks} &  \makecell{Open\\Drawer} &  \makecell{Close\\Jar}  &  \makecell{Stack\\Cups} &  \makecell{Sweep\\Dirt} &  \makecell{Slide\\Block} &  \makecell{Screw\\Bulb} &  \makecell{Put in\\Board} &  Avg. \\
    \midrule 
    CodeQwen-1.5-7B & \ding{55}  & 0 & 0 & 0 & 0 & 0 & 0 & 0 & 4 & 0 & 0.4\\
    CodeQwen-1.5-7B & \ding{51} & 68 & \bf 48 & \bf 68 & 44 & \bf 4 & \bf 48 & 60 & \bf 32 & \bf 12 & \bf 42.7\\
    DeepSeek-Coder-6.7B & \ding{51} & \bf 72 & 32 & \bf 68 & \bf 48 & \bf 0 & 24 & \bf 84 & \bf 32 & \bf 12 & 41.3\\
    \bottomrule
    \end{tabular}
\end{table*}

\begin{table*}[t]
    \centering
    \caption{Detailed success rate (\%) of ablation study on the LIBERO tasks. }
    \label{tab:ablation_libero}
    \small
    \setlength{\tabcolsep}{3pt} 
    \begin{tabular}{l|c|ccccccccccc}
    \toprule
    \bf LLM & \bf \ourdata{} &  \makecell{Turn on\\Stove} &  \makecell{Close\\Cabinet} &  \makecell{Put in\\Sauce} &  \makecell{Put in\\Butter}  &  \makecell{Put in\\Cheese} &  \makecell{Place\\Book} &  \makecell{Boil\\Water} &  \makecell{Identify\\Plate} &  Avg. \\
    \midrule 
    CodeQwen-1.5-7B & \ding{55}  & 0 & 43 & 0 & 0 & 13 & 0 & 0 & 0 & 7.0\\
    CodeQwen-1.5-7B & \ding{51} & \bf 97 & \bf 60 & \bf 67 & \bf 53 & \bf 63 & 43 & \bf 23 & 13 & \bf 52.4\\
    DeepSeek-Coder-6.7B & \ding{51} & \bf 97 & \bf 60 & 47 & \bf 53 & 60 & \bf 53 & 0 & \bf 20 & 48.8\\
    \bottomrule
    \end{tabular}
\end{table*}

\subsection{Tasks in LIBERO} 
\label{ap:task_libero}
In this section, we provide a detailed description of 8 tasks selected from the LIBERO-100 dataset. Each task is associated with a specific language instruction, with the task ID and corresponding instruction shown in Table~\ref{tab:task_detail}. The tasks \textit{"Turn on Stove"} and \textit{"Close Cabinet"} are taken from LIBERO-90, which focuses on testing atomic skills and environmental understanding. The remaining tasks are more complex, requiring multi-step execution, and are selected from LIBERO-10. These 8 tasks challenge \our{} to comprehend diverse visual environments and follow extended language instructions. As illustrated in Fig.~\ref{fig:libero_task}, the tasks encompass a wide range of robotic capabilities, including object selection, spatial reasoning, scene comprehension, and long-term execution.


\subsection{Tasks in Real-world Experiments}
\label{ap:real-world-setting}
\begin{figure}[!htbp]
    \centering
    \includegraphics[width=0.48\textwidth]{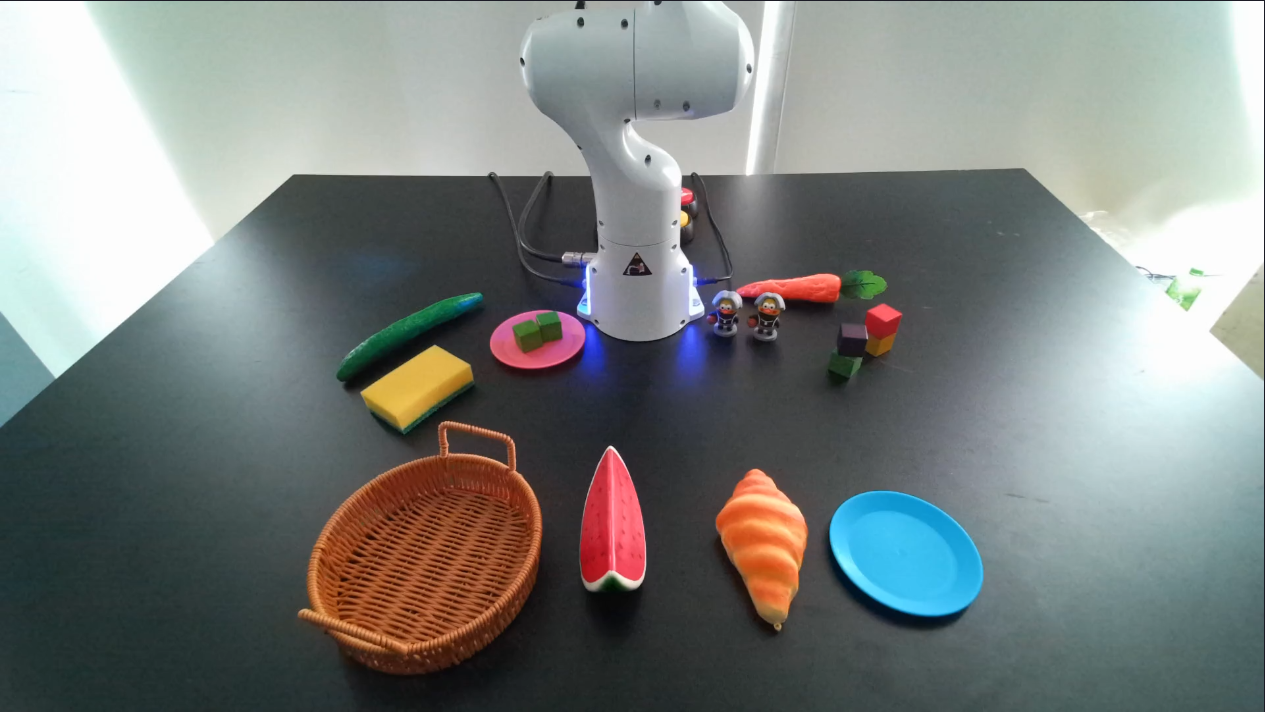}
    \caption{The setup for real-world experiments.}
    \label{fig:real_world_setup}
\end{figure}

\begin{figure*}[b]
    \centering
    \includegraphics[width=1\textwidth]{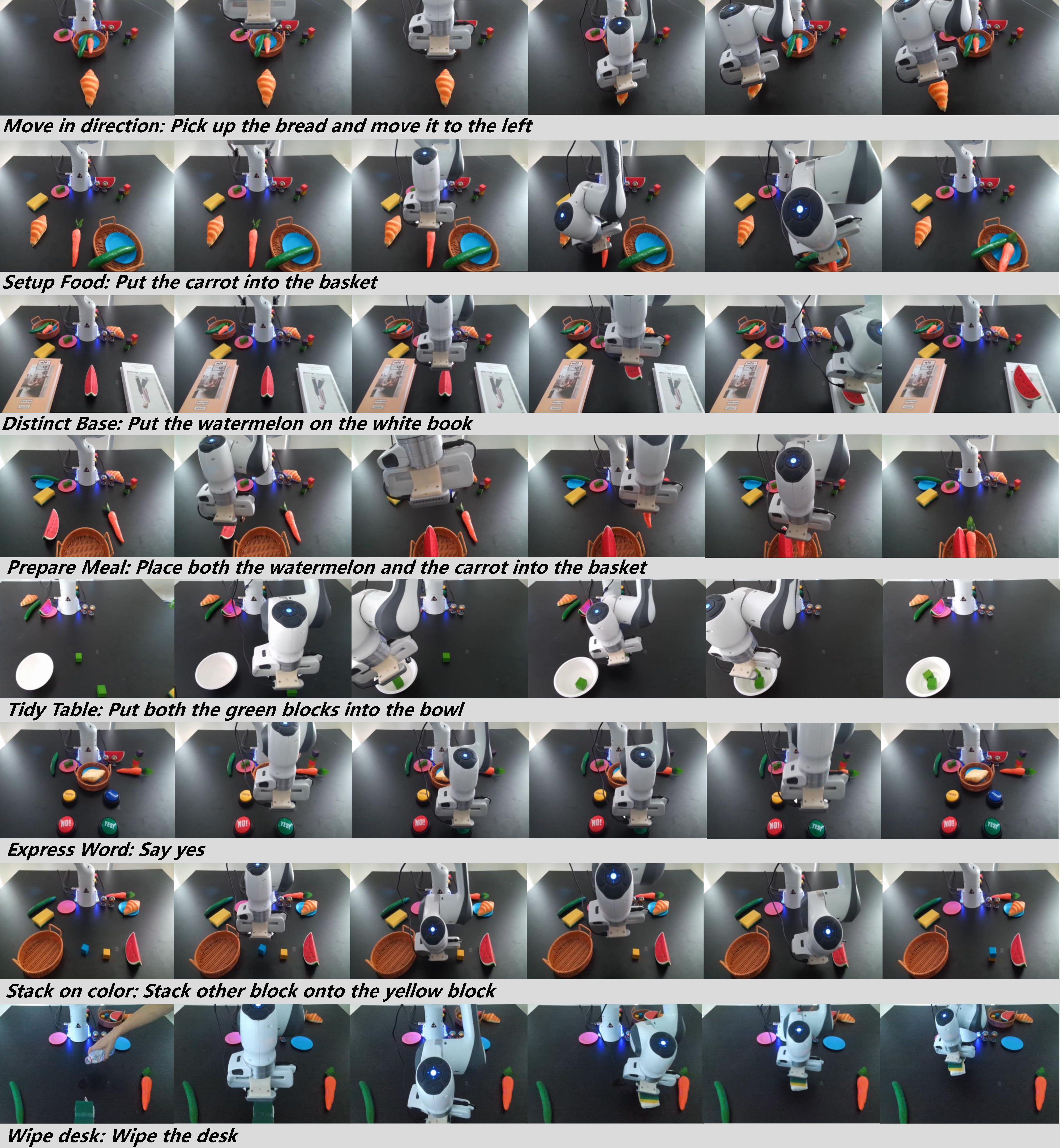}
    \caption{Illustration of \our{} on the real-world experiments.}
    \label{fig:real_world_task}
\end{figure*}

The real-world experiments are implemented on a Franka Emika Panda robotic arm with a parallel jaw gripper, as shown in Figure~\ref{fig:real_world_setup}. We use an Intel RealSense D435i camera to provide RGB-D input signals under the camera-in-hand setting. Easy-handeye ROS package is used to calibrate the extrinsics of the camera frame with respect to the robot base frame. For robot control, we use the open-source frankapy package to send real-time position-control commands to robot after receiving the control signals from \our{}. During test time for each task, natural language instructions, extrinsic matrix, intrinsic matrix, current environment observation in the form of RGB-D image, and the low dimensional state of the robot are prepared for \our{} to generate corresponding 6-DOF action trajectories. Examples of all 8 real-world tasks with natural language instructions are illustrated in Fig.~\ref{fig:real_world_task}, ranging from short-horizon to long-horizon tasks, as well as tasks that require visual comprehension.

\subsection{Additional Experiment Results}
\label{ap:exp}
We provide detailed results of our ablation study on two simulation platforms in Table~\ref{tab:ablation_libero} and Table~\ref{tab:ablation_rlbench}. As shown in the first two rows, the results on tow simulation platforms improved significantly after trained with \ourdata{} runtime data, which indicates the effectiveness of 115k visual-aligned code data collected from video demonstrations. As for the choice of base LLM, the version of RoboPro trained
with CodeQwen-1.5-7B-Chat consistently outperforms the
one trained with DeepSeek-Coder-6.7B-Instruct across both
manipulation tasks, which demonstrate that a better choice of code-domain LLM can consequently improve performance on robotic tasks.

\subsection{Prompts and API Implementations}
The API libraries provide interface for code generation, enabling low-level execution on different embodiments and tasks. As shown in Listing~\ref{lst:full}, the provided APIs can be divided into perception modules and action modules. The perception modules are primarily implemented for grounding of object positions and physical properties. For operations involving objects with similar properties, such as \textit{"stacking identical blocks"}, we provide the output in the form of a list of candidate objects. The action modules include direct motor execution via ROS, such as \texttt{open\_gripper()}, \texttt{rotate()}, and \texttt{move\_to\_pose()} with 7-DoF action inputs. Additionally, they provide motion planning APIs for generating trajectories around joint axes or predefined sequential paths to support skill implementations. For each of the APIs, detailed explanations and definitions of functions
and skills are provided in the prompt, which has proven to be a more effective approach for in-context learning
of policy code generation. We also make an effort to avoid hard-coding in the prompts, and use parameterized preferences and output of other APIs as input for grounding and motion generation, which makes the system flexible and adaptive to different scenes and objects. As mentioned in Sec.~\ref{exp:api}
, we discussed the influence of API Renaming and API Refactoring to the performance of \our{}, where variations of API implementations and prompts are depicted in Listing~\ref{lst:rename} and Listing~\ref{lst:reformat}
.

\label{ap:prompt}
\begin{figure*}[!htbp] 
    
\lstset{caption={An example of a full prompt provided for code generation models}, label=lst1}
\label{lst:full}
\begin{lstlisting}[language=Python]
"""You're a vision language model controlling a gripper to complete manipulation tasks. Combine the images you see with the text instructions to generate detailed and workable code for the current scene.
You have access to the following tools:
"""
------------
import numpy as np
import torch
import math

#Perception Modules
def get_obj_bbox(description: str)->list[bbox]:
"""get the 2D boundingbox of all objects match description. When it comes to the specific parts or orientation of objects, the description should be detailed. Like 'handle of microwave', 'left side of shelf'.
Return: list[bbox: np.ndarray]"""

def get_best_grasp_pos(grasp_bbox: bbox):
"""get best grasp pose to grasp specific object.
Return: grasp_pose: Pose"""

def get_place_pos(holder_bbox: bbox):
"""Predict the place pose for an object relative to a holder
Args: holder_bbox: bbox of target region of the holder.
Return: place_pose: Pose"""

def get_joint_axis(joint_object_name: str):
"""Get the joint direction of an object 
Args: joint_object_name: the name of object have joint axis.
Return: joint_axis: np.ndarray"""

def generate_joint_path(joint_axis: np.ndarray, open: bool):
"""Generate a gripper path of poses around the joint. open is True when need open container around joint, False when close container.
Return: path: list[Pose]
"""

def generate_slide_path(target: Optional[str] = None, direction: Optional[np.ndarray] = None):
"""Generate path of poses to slide or push object to target or in specific direction. 
Args: 
target: The target location. If provided, 'direction' must be None. 
direction: The direction vector to slide the object along. If provided, 'target' must be None.
Return: path: list[Pose]
"""

def generate_sweep_path(object: Optional[str] = None, target: Optional[str] = None, direction: Optional[np.ndarray] = None):
"""This function is designed to generate movement paths for sweeping actions using tools such as sweepers, brooms. Grasp the tool before sweeping.
Args: 
object: The object to be swept. If set to None, the function will perform a general sweeping. 
target: The target area or location to sweep towards. If provided, 'direction' must be None. 
direction: The direction vector for the sweeping motion. If provided, 'target' must be None.
Return: path: list[Pose]
"""

def generate_wipe_path(region: str):
"""This function is designed to generate movement paths for wiping actions using tools such as towel, sponge. Grasp the tool before wiping.
Args:
region (str): region to be wiped or cleaned.
Return: path: list[Pose]
"""

def generate_pour_path(grasped object: str, target: str):
"""Generate gripper path of poses to pour liquid in grasped object to target.
Return: path: list[Pose]
"""

def generate_press_pose(bbox):
"""Get best pose to press or push buttons."""

#Action Modules
def move_to_pose(Pose):
"""Move the gripper to pose."""

def move_in_direction(direction: np.ndarray, distance: float):
"""Move the gripper in the given direction in a straight line by certain distance.
"""

def follow_way(path: List[Pose]):
"""Move the gripper to follow a path of poses."""

def rotate(angle: float)
"""Rotate the gripper clockwise at certain degree while maintaining the original position."""

def open_gripper():
"""Open the gripper to release the object, no args"""

def close_gripper():
"""Close the gripper to grasp object, no args. Move to best grasp pose before close gripper."""

---------------
Rules you have to follow:
#Directions: right: [0,1,0], left: [0,-1,0], upward or lift object: [0,0,1], forward or move away: [1,0,0]
#Please solve the following instruction step-by-step. 
#You should ONLY implement the main() function and output in the Python-code style. Except the code block, output fewer lines.
---------------
Begin to excecute the task:
#Instruction: 
\end{lstlisting}

\end{figure*}

\begin{figure*}[!htbp] 

\lstset{caption={An example of a full prompt in \texttt{\our{}} with API renaming}, label=lst2}
\label{lst:rename}
\begin{lstlisting}[language=Python]
"""You're a vision language model controlling a gripper to complete manipulation tasks. Combine the images you see with the text instructions to generate detailed and workable code for the current scene.
You have access to the following tools:
"""
------------
import numpy as np
import torch
import math

#Perception Modules
def detect_bbox(description: str)->list[bbox]:
"""get the 2D boundingbox of all objects match description. When it comes to the specific parts or orientation of objects, the description should be detailed. Like 'handle of microwave', 'left side of shelf'.
Return: list[bbox: np.ndarray]"""

def generate_obj_grasp_pos(grasp_bbox: bbox):
"""get best grasp pose to grasp specific object.
Return: grasp_pose: Pose"""

def best_place_locator(holder_bbox: bbox):
"""Predict the place pose for an object relative to a holder
Args: holder_bbox: bbox of target region of the holder.
Return: place_pose: Pose"""

def find_axis_of_joint(joint_object_name: str):
"""Get the joint direction of an object 
Args: joint_object_name: the name of object have joint axis.
Return: joint_axis: np.ndarray"""

def map_joint_path(joint_axis: np.ndarray, open: bool):
"""Generate a gripper path of poses around the joint. open is True when need open container around joint, False when close container.
Return: path: list[Pose]
"""

def build_slide_path(target: Optional[str] = None, direction: Optional[np.ndarray] = None):
"""Generate path of poses to slide or push object to target or in specific direction. 
Args: 
target: The target location. If provided, 'direction' must be None. 
direction: The direction vector to slide the object along. If provided, 'target' must be None.
Return: path: list[Pose]
"""

def sweep_motion_path(object: Optional[str] = None, target: Optional[str] = None, direction: Optional[np.ndarray] = None):
"""This function is designed to generate movement paths for sweeping actions using tools such as sweepers, brooms. Grasp the tool before sweeping.
Args: 
object: The object to be swept. If set to None, the function will perform a general sweeping. 
target: The target area or location to sweep towards. If provided, 'direction' must be None. 
direction: The direction vector for the sweeping motion. If provided, 'target' must be None.
Return: path: list[Pose]
"""

def create_wipe_path(region: str):
"""This function is designed to generate movement paths for wiping actions using tools such as towel, sponge. Grasp the tool before wiping.
Args:
region (str): region to be wiped or cleaned.
Return: path: list[Pose]
"""

def pour_path_mapper(grasped object: str, target: str):
"""Generate gripper path of poses to pour liquid in grasped object to target.
Return: path: list[Pose]
"""

def best_press_pos(bbox):
"""Get best pose to press or push buttons."""

#Action Modules
def relocate_to_pose(Pose):
"""Move the gripper to pose."""

def reach_in_direction(direction: np.ndarray, distance: float):
"""Move the gripper in the given direction in a straight line by certain distance.
"""

def follow_path(path: List[Pose]):
"""Move the gripper to follow a path of poses."""

def spin_gripper(angle: float)
"""Rotate the gripper clockwise at certain degree while maintaining the original position."""

def open_claw():
"""Open the gripper to release the object, no args"""

def clamp_gripper():
"""Close the gripper to grasp object, no args. Move to best grasp pose before close gripper."""

---------------
Rules you have to follow:
#Directions: right: [0,1,0], left: [0,-1,0], upward or lift object: [0,0,1], forward or move away: [1,0,0]
#Please solve the following instruction step-by-step. 
#You should ONLY implement the main() function and output in the Python-code style. Except the code block, output fewer lines.
---------------
Begin to excecute the task:
#Instruction: 
\end{lstlisting}

\end{figure*}

\begin{figure*}[!htbp] 

\lstset{caption={An example of a full prompt for \texttt{\our{}} with API refactoring}, label=lst3, float=H}
\label{lst:reformat}
\begin{lstlisting}[language=Python]
"""You're a vision language model controlling a gripper to complete manipulation tasks. Combine the images you see with the text instructions to generate detailed and workable code for the current scene.
You have access to the following tools:
"""
------------
import numpy as np
import torch
import math

#Perception APIs
def get_obj_bbox(description: str) -> list[np.ndarray]:
"""Get the 2D bounding box of all objects that match the description. The description should be detailed when it comes to specific parts or orientations of objects, such as 'handle of microwave' or 'left side of shelf'.
Args:description (str): The description of the objects to find.
Returns:list[np.ndarray]: A list of bounding boxes for the objects matching the description."""

def get_joint_axis(joint_object_bbox: np.ndarray):
"""Get the joint direction of an object 
Args: joint_object_name: the name of object have joint axis.
Return: joint_axis: np.ndarray"""

#Control APIs
def get_best_grasp_pos(grasp_bbox: np.ndarray):
"""Calculate the best grasp pose to grasp a specific object.
Parameters: grasp_bbox (np.ndarray): The bounding box of the object to grasp.
Return: Pose: The best grasp pose for the given object."""

def get_place_pos(holder_bbox: np.ndarray):
"""Predict the place pose for an object relative to a holder.
Parameters: holder_bbox (np.ndarray): The bounding box of the target region of the holder.
Return: Pose: The predicted place pose for the given object."""

def generate_joint_path(joint_axis: np.ndarray, open: bool) -> list[Pose]:
"""Generate a gripper path of poses around the joint.
Parameters: joint_axis (np.ndarray): The axis of the joint. open (bool): True if the container needs to be opened around the joint, False if it needs to be closed.
Returns: list[Pose]: The generated path of poses around the joint."""

def generate_slide_path(target_bbox: np.ndarray) -> List[Pose]:
"""Generate a path of poses to slide or push an object to a target or in a specific direction.
Parameters: target_bbox (np.ndarray): bbox of the target location. 
Returns: List[Pose]: The generated path of poses."""

def generate_sweep_path(target_bbox: np.ndarray) -> List[Pose]:
"""Generate movement paths for sweeping actions using tools such as sweepers or brooms. Grasp the tool before sweeping.
Parameters: target_bbox (np.ndarray): The target area or location to sweep towards. 
Returns: List[Pose]: The generated path of poses for the sweeping action."""

def generate_wipe_path(region_bbox: np.ndarray) -> List[Pose]:
"""Generate movement paths for wiping actions using tools such as towels or sponges. Grasp the tool before wiping.
Parameters: region_bbox (np.ndarray): The region to be wiped or cleaned.
Return: List[Pose]: The generated path of poses for the wiping action."""

def generate_pour_path(grasped_object: str, target_bbox: np.ndarray) -> List[Pose]:
"""Generate a gripper path of poses to pour liquid from a grasped object to a target.
Parameters: grasped_object (str): The object being grasped that contains the liquid. target_bbox (np.ndarray): The bounding box of the target area where the liquid will be poured.
Returns: List[Pose]: The generated path of poses for the pouring action."""

def generate_press_pose(bbox: np.ndarray) -> Pose:
"""Get the best pose to press or push buttons.
Parameters: bbox (BBox): The bounding box of the button or area to be pressed.
Return: Pose: The best pose for pressing or pushing the button."""

def move_to_pose(pose: Pose):
"""Move the gripper to the specified pose.
Parameters: pose (Pose): The target pose to move the gripper to."""

def move_in_direction(direction: np.ndarray, distance: float):
"""Move the gripper in the given direction in a straight line by a certain distance.
Parameters: direction (np.ndarray): The direction vector to move the gripper along. distance (float): The distance to move the gripper."""

def follow_way(path: List[Pose]) -> None:
"""Move the gripper to follow a path of poses.
Parameters: path (List[Pose]): The list of poses that defines the path to follow."""

def rotate(angle: float) -> None:
""" Rotate the gripper clockwise by a certain angle while maintaining the original position.
Parameters: angle (float): The angle in degrees to rotate the gripper."""

def open_gripper() -> None:
"""Open the gripper to release the object.
Parameters: None
Returns: None"""

def close_gripper() -> None:
"""Close the gripper to grasp an object. Move to the best grasp pose before closing the gripper.
Parameters: None
Returns: None"""

---------------
Rules you have to follow:
#Directions: right: [0,1,0], left: [0,-1,0], upward or lift object: [0,0,1], forward or move away: [1,0,0]
#Please solve the following instruction step-by-step. 
#You should ONLY implement the main() function and output in the Python-code style. Except the code block, output fewer lines.
---------------
Begin to excecute the task:
#Instruction: 
\end{lstlisting}
\end{figure*}

\end{document}